\documentclass[conference]{IEEEtran}
\IEEEoverridecommandlockouts
\usepackage{cite}
\usepackage{amsmath,amssymb,amsfonts}
\usepackage{algorithmic}
\usepackage{graphicx}
\usepackage{textcomp}
\usepackage{xcolor}

\def\BibTeX{{\rm B\kern-.05em{\sc i\kern-.025em b}\kern-.08em
    T\kern-.1667em\lower.7ex\hbox{E}\kern-.125emX}}
\begin{document}

\title{Combining Neural Network Models for Blood Cell Classification\\
}

\author{\IEEEauthorblockN{1\textsuperscript{st} Indraneel Ghosh}
\IEEEauthorblockA{\textit{Department of Computer Science and Information Systems} \\
\textit{Department of Biological Sciences} \\
\textit{Birla Institute of Technology and Science}\\
Pilani, Rajasthan, India \\
f2016938@pilani.bits-pilani.ac.in}
\and
\IEEEauthorblockN{2\textsuperscript{nd} Siddhant Kundu}
\IEEEauthorblockA{\textit{Department of Computer Science and Information Systems} \\
\textit{Birla Institute of Technology and Science}\\
Pilani, Rajasthan, India \\
f2016055@pilani.bits-pilani.ac.in}
}

\maketitle
\begin{abstract}
The objective of the study is to evaluate the efficiency of a multi layer neural network models  built by combining Recurrent Neural Network(RNN) and Convolutional Neural Network(CNN) for solving the problem of classifying of different kind of White Blood Cells. This can have applications in the pharmaceutical and healthcare industry for automating the  analysis of blood tests and other processes requiring identifying the nature of blood cells in a given image sample.  It can also be used in diagnosis of various blood related diseases in patients. 
\end{abstract}


\section{Introduction}
White Blood Cells are an important component of our immune system. The concentration of various WBCs play an important role in determining the health of our body. Automating the process of detection of these blood cells accurately can help expedite various medical processes. Each class of White Blood Cell is unique. They have differences in texture, color size and morphology. Our approach attempts to classify the WBCs based on the latent features of their images.

\section{Methods}
The two main segments of the architecture of our proposed neural network are a Convolutional Neural Network and a Recurrent Neural Network, both trained using the same image data.

\subsection{Convolutional Neural Networks}
A Convolutional Neural Network (CNN) is a type of neural network containing cells that extract features from an input by moving over it with a small window, called a kernel. The kernel moves over the entire input, and the portion of the image captured in the kernel window is checked for features corresponding to the one the cell has learned to detect. Applied sequentially, Convolutional Neural Networks are capable of extracting both high-level and low-level features. They are typically applied to images, where the usage of a simple feedforward network would make the hidden layers unnecessarily large and computationally expensive, and also prone to overfitting.

\subsection{Recurrent Neural Networks}
A Recurrent Neural Network (RNN) is a variation of standard feedforward networks where the output of a layer is dependent not only on the current input, but also the set of inputs that has come before. This is useful for sequence detection and generation. They provide a significant advantage when the inputs obtained before can be used to predict what kind of output comes later.

\subsection{The Advantages of LSTMs}
The standard RNN architecture uses a hidden state $h_{t}$ in addition to an input and an output. Its equations can be represented as:
\begin{equation}
  \label{eq:rnn_hidden}
  h_{t} = \tanh(W_{hh}h_{t-1}+W_{xh}X_{t})
\end{equation}
\begin{equation}
  \label{eq:rnn_output}
  y_{t} = \sigma(W_{hy}h_{t})
\end{equation}
One of the biggest challenges in the model that we have built is handling long-term dependencies. Theoretically, an RNN is capable of handling long term dependencies with careful parameter selection. However, it has been noted that in practice this is not true\cite{b5}. \\
Long Short Term Memory Networks (LSTMs) were created to handle these long term dependencies specifically\cite{b6}. As a result, we used LSTM while building our model.
Their structure may be described as:
\begin{equation}
  \label{eq:lstm_forget}
  f_{t} = \sigma(W_{f}[h_{t-1},X_{t}]+b_{f})
\end{equation}
\begin{equation}
  \label{eq:lstm_input}
  i_{t} = \sigma(W_{i}[h_{t-1},X_{t}]+b_{i})
\end{equation}
\begin{equation}
  \label{eq:lstm_new}
  \hat{C_{t}} = \tanh(W_{C}[h_{t-1},X_{t}]+b_{C})
\end{equation}
\begin{equation}
  \label{eq:lstm_cell}
  C_{t} = f_{t}*C_{t-1} + i_{t}*\hat{C_{t}}
\end{equation}
\begin{equation}
  \label{eq:lstm_output}
  o_{t} = \sigma(W_{o}[h_{t-1},X_{t}]+b_{o})
\end{equation}
\begin{equation}
  \label{eq:lstm_hidden}
  h_{t} = o_{t}*\tanh(C_{t})
\end{equation}

Here, $X_{t}$ is the input vector at timestamp t, which denotes the t\textsuperscript{th} item in the time series form of the input. $h_{t}$ denotes the hidden state of the LSTM, which works in roughly the same manner as for a regular RNN, and is used for storing information about previous inputs. $f_{t}$ denotes the forget gate - i.e. the part of the previous hidden state that needs to be forgotten for the current input. $i_{t}$ denotes the input gate, which decides which parts of the input vector should be stored in the next hidden state. $C_{t}$ denotes the cell state at t, which is used for output generation, and $\hat{C_{t}}$ denotes the set of values to be updated at timestamp t. Finally, $o_{t}$ denotes the output of the LSTM.

Our work combines the CNN layer, that classifies images based on a hierarchy of features ranging from straight lines and curves to specific nuclear shapes, and the RNN layer, that uses the fact that we are classifying white blood cell images that will contain certain fixed low level features leading to some more fixed high level features, in sequence. To obtain a faster, optimized model we make use of  NVIDIA CUDA\textregistered \\ Deep Neural Network library (cuDNN)  which significantly reduces the training time of the model.

\subsection{Dataset}
The dataset obtained from Kaggle contains 12,500 augmented images of blood cells in JPEG format with the accompanying cell type labels. The cell types are Eosinophil, Lymphocyte, Monocyte, and Neutrophil. There are approximately 3,000 images for each of the four cell types. This dataset is accompanied by an additional dataset containing the original pre-augmented 410 images as well as two sub-type labels and also bounding boxes for labeling (JPEG+XML) each of the cells in these images. It also contains 2500 augmented images as well as four additional subtype labels (JPEG+ CSV). There are approximately 2,000 augmented images for each of the four class as compared to 88, 33,21 and 207 of the original images.\cite{b3}
\begin{figure}[htbp]
\centerline{\includegraphics[width=5.5cm,height=5.5cm,keepaspectratio]{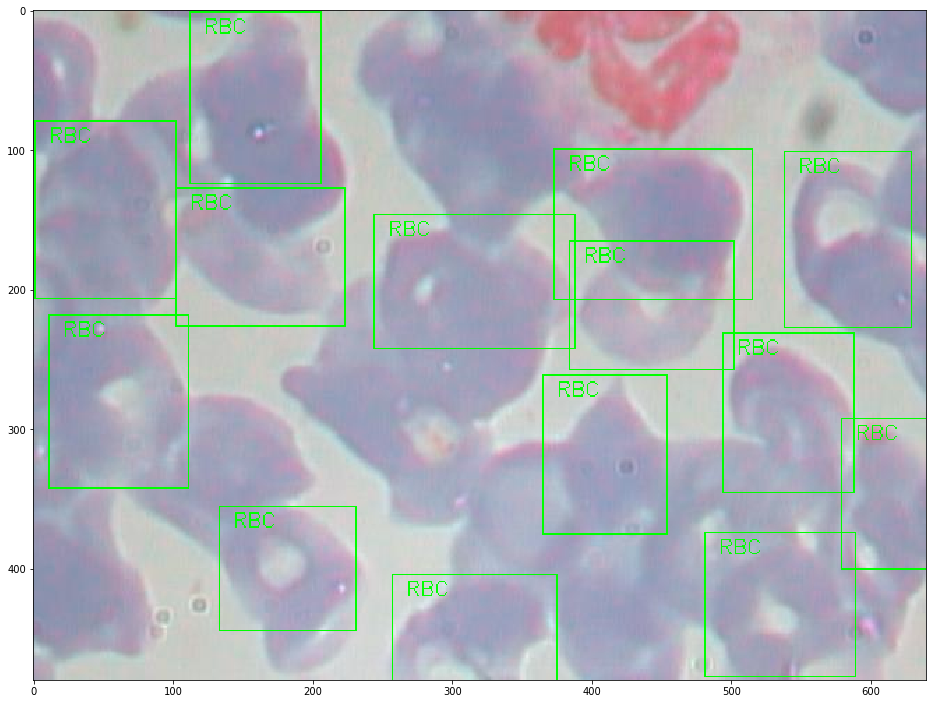}}
\caption{A Labelled Blood Cell Image }
\label{fig}
\end{figure}

\subsection{Data Pre-Processing}
Since the dataset we have is small, we decided to augment the images by rotation, reflection about the horizontal axis and shifting both horizontally and vertically. We must ensure that the computation time without losing too much accuracy. Hence, the size of the input image was reduced  to $80\times60$. Scaling transformations are not applied because the correct identification of the type of cell depends on the size of the nuclei.

The four cell types have been transformed into a 4 dimensional vector with one-hot encoding - i.e., all components are zero, except the one corresponding to the appropriate class. For example, the cell type 'NEUTROPHIL' may be encoded as [0,0,1,0], and 'MONOCYTE' may be encoded as [1,0,0,0].  This transformation allows us to use the softmax loss function\cite{b4}, defined as:
\begin{equation}
  \label{eq:softmax}
  y_k = \frac{\exp(\phi_k)}{\sum^{c}_j \exp(\phi_j)},
\end{equation}

\subsection{Model}
Our Model consists of five parts - the Input Layer, the CNN Layer, the RNN Layer, the Merge Layer and the Output Layer - as detailed below.

\subsubsection{Input Layer}
This is the simplest layer of the network. It takes the image data as an input, and converts it into a tensor of the appropriate size - $img\_rows\times image\_columns\times 3$ (one for each primary color).

\subsubsection{CNN Layer}
This is the segment consisting of convolutional cells that scan the image for present features. In our model, it contains 4 layers. The first layer is a convolutional layer that consists of 32 cells, and the second layer is also a convolutional layer that consists of 64 cells. Each layer uses a $3\times3$ kernel. It is followed by pooling to reduce the size of the output, a dropout layer to reduce overfitting and a layer that flattens the output to 2D.

\subsubsection{RNN Layer}
This layer consists of LSTMs that learn to detect recurring features in the images.
In our model, the image is converted to grayscale before being passed to the RNN, to reduce overfitting due to learning color patterns. This is likely to happen since the dataset is quite small (even after augmentations) and RNNs do not deal with overfitting due to color in images as well as CNNs due to their similarities with standard feedforward networks. Our model has 2 LSTM layers. Each LSTM layer uses 64 LSTMs, and is followed by a dropout layer to reduce overfitting.

\subsubsection{Merge Layer}

The merge layer takes the CNN layer's output and the RNN layer's output and returns the element-wise product of the two. Both layers were constructed to have 64 neurons at the time of merging. Therefore, we get the combination of the CNN and RNN parts. A 128 cell layer with ReLU activation follows this for another round of processing, followed by a final dropout layer, feeding into the output layer.

\subsubsection{Output Layer}

The output layer has 128 cells, and applies the softmax function to evaluate which class the given input image is most likely to fall into, as defined by the input set, according to equation
\eqref{eq:softmax}. Since the input set only has single-class data, we decided to use the softmax function to improve our model's classification power.

\subsection{Training}

For training the network, we decided to use the predefined split in the dataset. Both the branches of the network, the CNN and the RNN, are trained simultaneously. The loss function used is categorical cross-entropy, since we have a four-category classification problem. The optimizer used is Adadelta, since it allows for adaptive learning rate based on previous gradients.

For both the four-way and the two-way classification problems, the model is trained for 70 epochs, with an initial learning rate of 1.0 and a batch size of 32.

\section{Experiments and Results}
For comparative purposes, we tested a CNN-only solution to the problem, under the same conditions as the hybrid solution. A table comparing the two is shown below:
\begin{table}[htbp]
\caption{Comparison of Accuracies of the Different Architectures}
\begin{center}
\begin{tabular}{|c|c|c|}
\hline
\textbf{Architectures}&\multicolumn{2}{|c|}{\textbf{Classification Accuracy}} \\
\cline{2-3} 
\textbf{} & \textbf{\textit{2 Way Classification}}& \textbf{\textit{4 Way Classification}} \\
\hline

CNN  & 93.76\% & 86.16\%  \\
\hline
CNN + RNN & 94.13\% & 87.29\%  \\
\hline
\end{tabular}
\label{tab1}
\end{center}
\end{table}

\section*{}

\begin{figure}[htbp]
\centerline{\includegraphics[width=9cm,height=9cm,keepaspectratio]{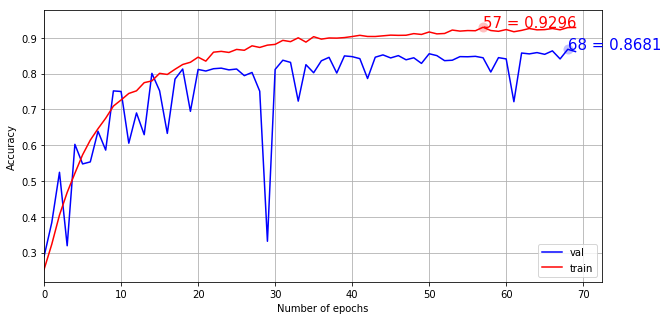}}
\caption{CNN only 4 way classification}
\label{fig}
\end{figure}

\begin{figure}[htbp]
\centerline{\includegraphics[width=9cm,height=9cm,keepaspectratio]{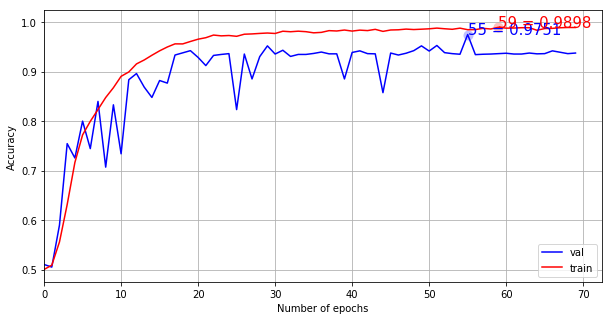}}
\caption{CNN only 2 way classification }
\label{fig}
\end{figure}
\begin{figure}[htbp]
\centerline{\includegraphics[width=9cm,height=9cm,keepaspectratio]{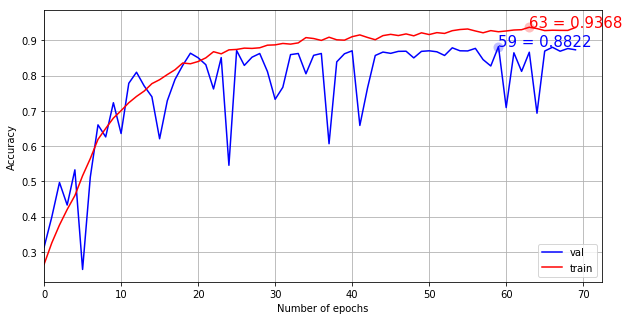}}
\caption{CNN +RNN 4 way classification }
\label{fig}
\end{figure}

\begin{figure}[htbp]
\centerline{\includegraphics[width=9cm,height=9cm,keepaspectratio]{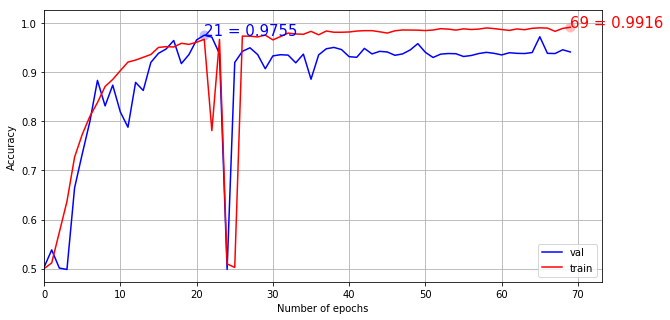}}
\caption{CNN+RNN 2 way classification }
\label{fig}
\end{figure}
\newpage
\section*{Conclusion}
We can see that the CNN-RNN hybrid architecture offers better results than the CNN-only network - for the four-way classification problem, approximately 8\% of the cases misclassified by the CNN-only solution are classified correctly by the CNN-RNN combination, assuming that the hybrid network does not misclassify the items that are correctly classified by the CNN-only network. For the two-way classification problem, approximately 6\% of the misclassified cases are correctly classified by the hybrid network. With the usage of the optimized cuDNN version of the LSTM layers, the training time is roughly the same as with the pure CNN network (approximately 13 seconds per epoch for CNN only, approximately 17 seconds per epoch on CNN+RNN). Thus, we can conclude that the usage of an RNN alongside a CNN leads to improved performance.

\section*{Acknowledgement}
The authors would like to thank the Department of Biological Sciences,BITS Pilani for providing us with an opportunity to present our work and fine-tuning our ideas.

\vspace{12pt}

\end{document}